\documentclass[runningheads]{llncs}
\usepackage[T1]{fontenc}
\usepackage{graphicx}
\usepackage{booktabs}
\usepackage{multirow}
\usepackage{array}
\usepackage{amsmath}
\usepackage{esvect}
\usepackage{url}
\usepackage{float}

\begin{document}

\title{Clustering of High-Dimensional Human Gene Score Data for Target Cell Classification}

\titlerunning{Target Cell Classification via Gene Scores}

\author{
  S. Das\inst{1} \and
  V. Madala\inst{2} \and
  P. Majumder\inst{2} \and 
  N. S. Sinha\inst{3}
}

\authorrunning{S. Das et al.} 

\institute{
  School of Mathematics, Indian Institute of Science Education and Research Thiruvananthapuram, Kerala, India 695551 \and
  School of Data Sciences, Indian Institute of Science Education and Research Thiruvananthapuram, Kerala, India 695551 \and
  School of Biological Sciences, Indian Institute of Science Education and Research Thiruvananthapuram, Kerala, India 695551
}

\maketitle

\begin{abstract}
The goal is to investigate whether the total human gene-score dataset and the selected human kinase-gene dataset exhibit the same number of clusters when analysed using the same set of 769 features. Two datasets were considered: the complete gene score dataset and a subset containing selected kinase genes. Two missing-data strategies were investigated. In Case I, rows containing missing gene-score values were removed. In Case II, rows with more than 20\% missing values were excluded and the other missing values were replaced by the corresponding column mean. For each case, the data were scaled and principal component analysis (PCA) was performed before clustering. Four clustering approaches were considered: K-Means, Hierarchical Clustering, Fuzzy C-Means, and Graph-based Clustering. Across both datasets and both missing-data treatments, the optimal number of clusters was found to be two. In the absence of an optimal number of clusters, the quality of clustering was assessed using the Silhouette score. Relatively high Silhouette scores were obtained in all cases.
\end{abstract}

\section{Introduction}
High-dimensional gene score data can contain a large number of features for each gene, making it useful to investigate the underlying structure of the data through clustering. Clustering is a widely used unsupervised learning approach for identifying structure in high-dimensional data, with partitioning, hierarchical, fuzzy, and graph-based approaches providing complementary perspectives on data organization \cite{jain2010,berkhin2006}. The present study focuses on determining whether the total human gene score data and a selected human kinase dataset form the same number of clusters when analysed using the same 769 features.

Two datasets were considered. The first, referred to as the mother or total gene score data, contains 19,814 genes with 769 features. The second contains selected kinase genes from the mother dataset. The analysis was performed under two different treatments of missing values and using multiple clustering methodologies.

The main objective is to examine whether both datasets exhibit a consistent clustering structure and, in particular, whether they form the same number of optimal clusters under the same feature space.

\section{Data and Methodology}

\subsection{Datasets}
Two datasets were considered:
\begin{itemize}
    \item \textbf{Mother data / total gene score data:} $19,814 \times 769$.
    \item \textbf{Human kinase data:} 518 kinase gene’s data selected from the mother dataset, represented by the same 769 features.

\end{itemize}

\subsection{Missing-Data Treatments}
Two cases were considered.

\textbf{Case I: Complete Data Analysis.}  If any missing gene-score value was present in a row, the entire row was removed. The final datasets consisted of 16,138 gene-score values and 418 kinases.

\textbf{Case II: Mean Imputation.} Rows containing more than 20\% missing values were excluded, following the missing-data handling strategy used in the project report; mean imputation is a standard approach for completing missing numerical observations, although its assumptions and limitations should be considered \cite{little2019}. In the remaining data, missing values were replaced by the corresponding column mean. The resulting datasets contained 18,073 gene-score observations and 498 kinase observations, both with 769 features.

\subsection{Pre-processing}
The data were scaled before clustering. Principal Component Analysis (PCA) was carried out on the scaled datasets to capture the major directions of variation in the data while retaining low-dimensional representations \cite{jolliffe2002}. The representations were used for clustering with the same set of 769 features.

\subsection{Clustering Methods}
Four clustering methods were considered.

\begin{itemize}
    \item \textbf{K-Means Clustering:} This method, originating from the classical partitioning literature \cite{macqueen1967,lloyd1982}, was selected because it can tackle large datasets efficiently. The output is readily interpretable because each observation is assigned to the nearest cluster centre. Since the clusters' number must be specified, the elbow method was used to find the optimal number of clusters.
    \item \textbf{Hierarchical Clustering:} This family of methods provides a hierarchical representation of the data through a dendrogram \cite{johnson1967}. The dendrogram can be used to identify an appropriate number of clusters without specifying the number in advance.
    \item \textbf{Fuzzy C-Means:} This method provides soft clustering, following the fuzzy clustering framework of Bezdek \cite{bezdek1981}, allowing observations to have membership in multiple clusters. This can be useful when observations do not strictly belong to a single cluster. The method was also selected because it can be robust to noisy observations and outliers.
    \item \textbf{Graph-based Clustering:} This method was considered because graph and spectral formulations can capture complex relationships and structures in data \cite{luxburg2007} and can account for noisy observations and outliers.
\end{itemize}

\subsection{Cluster Evaluation}

The number of clusters was assessed using the corresponding clustering procedures, and the elbow method was used for the K-Means procedure. The Silhouette score was also computed for each of the clustering procedures. Higher scores indicate that observations in a particular cluster are more similar to each other compared to the other clusters \cite{rousseeuw1987}. Generally, higher values of the Silhouette score indicate that the clustering structure is better defined.

\section{Results and Findings}
The results of the four clustering algorithms for both the human gene score data and kinase data are summarized for Case I and Case II in Tables~\ref{tab:case1} and~\ref{tab:case2}.

\begin{table}[H]
\caption{Results of various clustering algorithms under complete data analysis (Case I).}
\label{tab:case1}
\centering
\small
\begin{tabular}{lcccc}
\toprule
& \multicolumn{2}{c}{Gene Score Data ($16138\times769$)} & \multicolumn{2}{c}{Kinase Data ($418\times769$)}\\
\cmidrule(lr){2-3}\cmidrule(lr){4-5}
Clustering Method & No. clusters & Silhouette & No. clusters & Silhouette\\
\midrule
K-Means & 2 & 0.770 & 2 & 0.724\\
Hierarchical & 2 & 0.788 & 2 & 0.825\\
Fuzzy C-Means & 2 & 0.768 & 2 & 0.818\\
Graph-based & 2 & 0.633 & 2 & 0.817\\
\bottomrule
\end{tabular}
\end{table}

\begin{table}[H]
\caption{Results of various clustering algorithms under mean imputation (Case II).}
\label{tab:case2}
\centering
\small
\begin{tabular}{lcccc}
\toprule
& \multicolumn{2}{c}{Gene Score Data ($18073\times769$)} & \multicolumn{2}{c}{Kinase Data ($498\times769$)}\\
\cmidrule(lr){2-3}\cmidrule(lr){4-5}
Clustering Method & No. clusters & Silhouette & No. clusters & Silhouette\\
\midrule
K-Means & 2 & 0.773 & 2 & 0.805\\
Hierarchical & 2 & 0.788 & 2 & 0.825\\
Fuzzy C-Means & 2 & 0.771 & 2 & 0.807\\
Graph-based & 2 & 0.633 & 2 & 0.795\\
\bottomrule
\end{tabular}
\end{table}

For both missing data treatments and both the datasets, the number of clusters was optimized to two. Also, in all cases, the values of the Silhouette score were relatively high. Therefore, the observations in a particular cluster were similar to each other, and the differences were observed with respect to other clusters.

\section{Clustering Visualizations}

\subsection{K-Means Clustering}

\subsubsection{Complete Data Analysis}
\begin{figure}[H]
\centering
\includegraphics[width=0.35\textwidth]{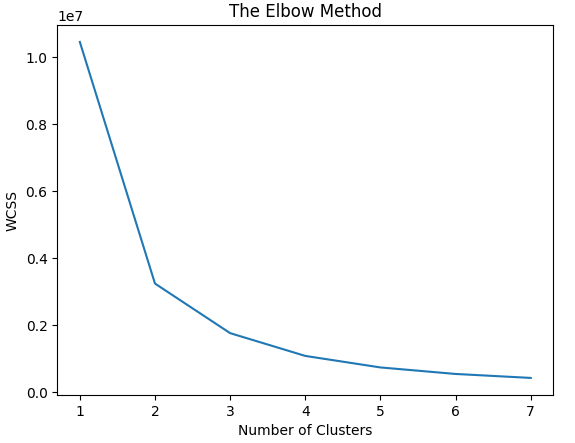}\hfill
\includegraphics[width=0.65\textwidth]{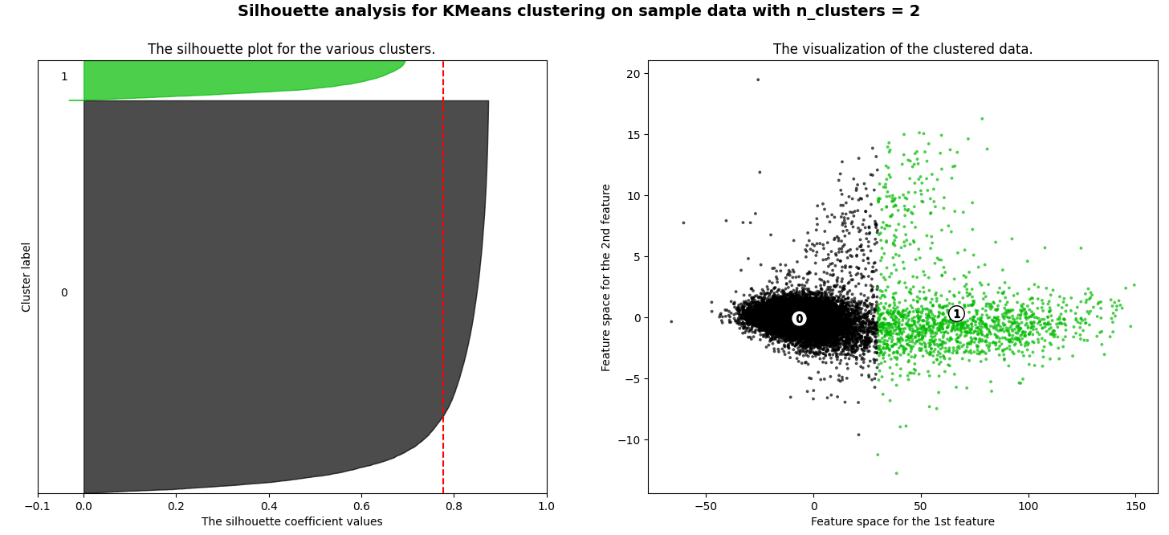}
\caption{K-Means clustering for the mother dataset under complete data analysis: elbow method and silhouette/cluster visualization.}
\label{fig:kmeans-complete-mother}
\end{figure}

Figure~\ref{fig:kmeans-complete-mother} presents the K-Means analysis for the mother dataset, including the elbow analysis and the silhouette/cluster visualization.

\begin{figure}[H]
\centering
\includegraphics[width=0.37\textwidth]{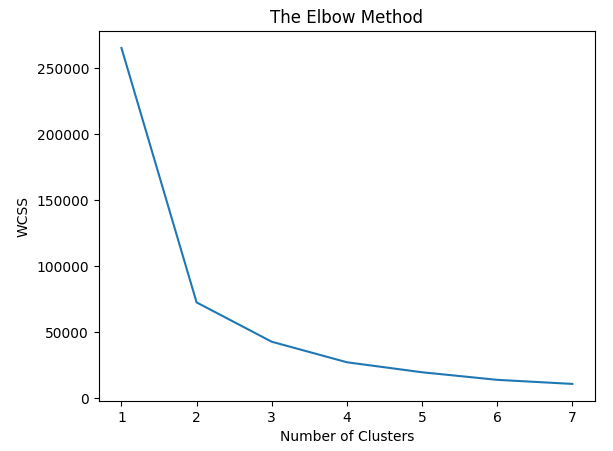}\hfill
\includegraphics[width=0.63\textwidth]{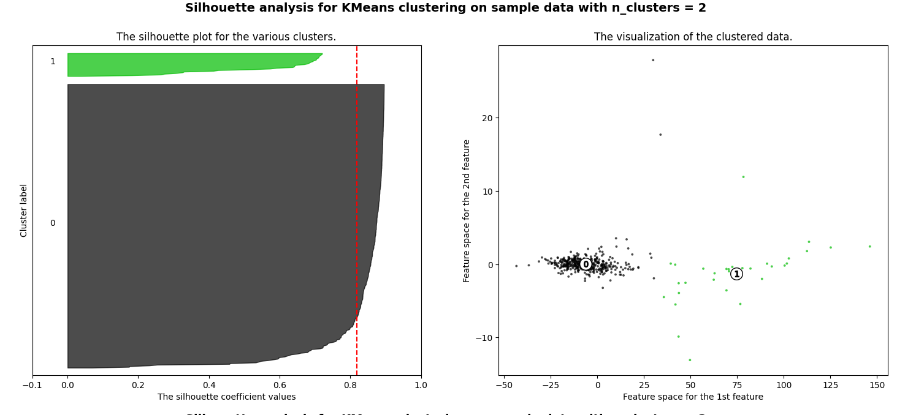}
\caption{K-Means clustering for the kinase dataset under complete data analysis.}
\label{fig:kmeans-complete-kinase}
\end{figure}

Figure~\ref{fig:kmeans-complete-kinase} shows the corresponding K-Means analysis for the kinase dataset.

\subsubsection{20\% Missingness Followed by Column-Mean Imputation}
\begin{figure}[H]
\centering
\includegraphics[width=0.64\textwidth]{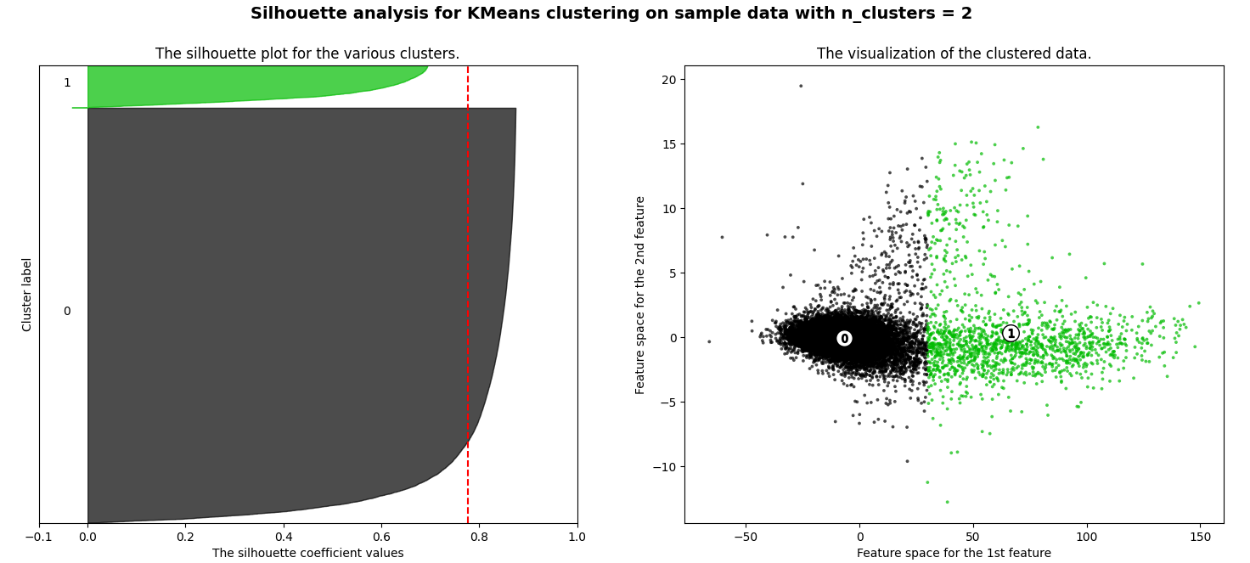}\hfill
\includegraphics[width=0.36\textwidth]{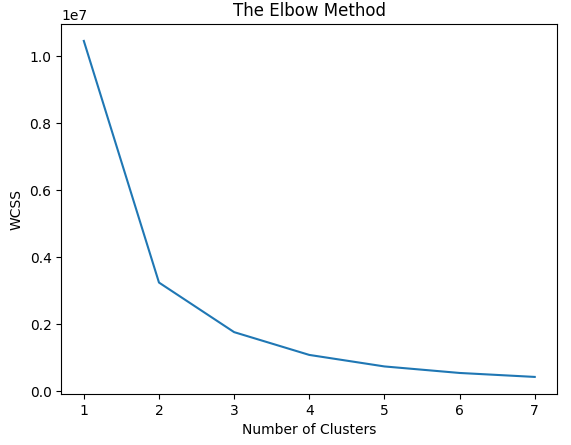}
\caption{K-Means clustering for the mother dataset under Case II.}
\label{fig:kmeans-imputed-mother}
\end{figure}

\begin{figure}[H]
\centering
\includegraphics[width=0.37\textwidth]{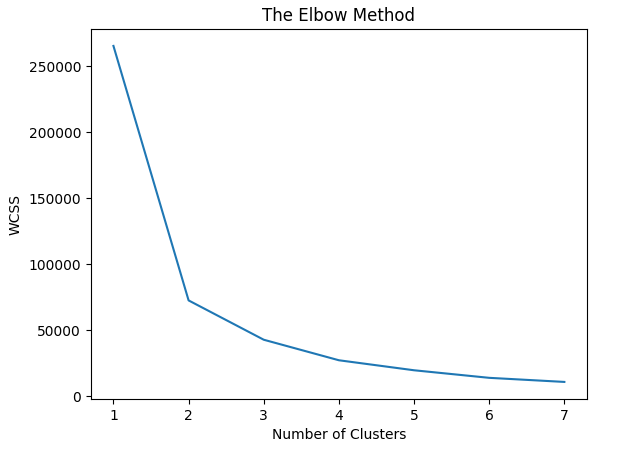}\hfill
\includegraphics[width=0.63\textwidth]{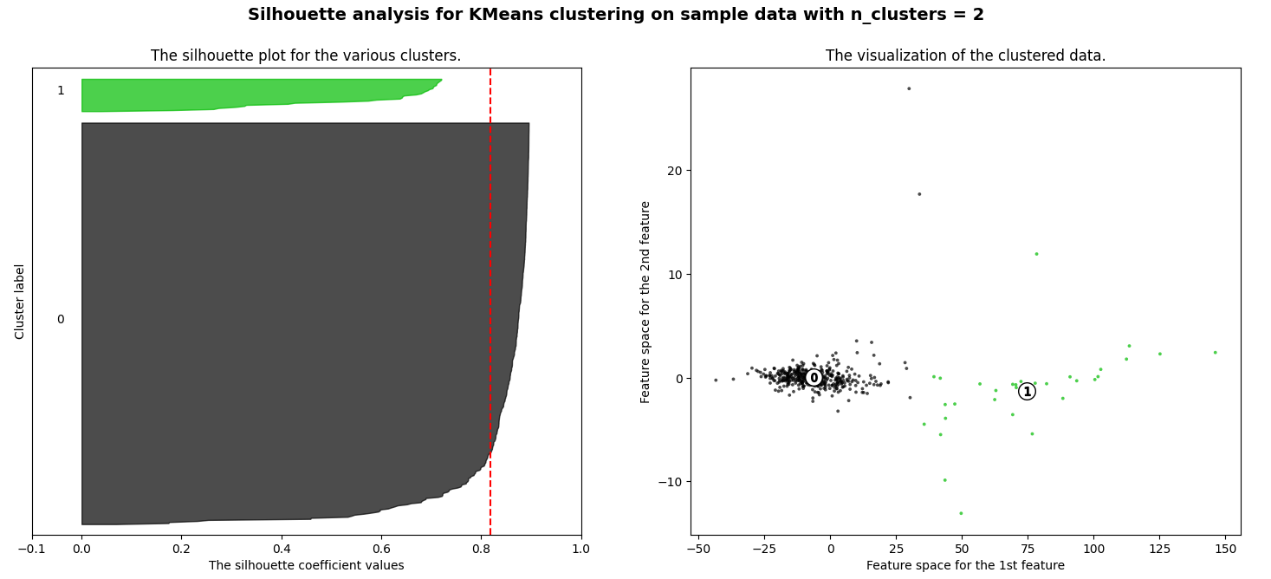}
\caption{K-Means clustering for the kinase dataset under Case II.}
\label{fig:kmeans-imputed-kinase}
\end{figure}

Figures~\ref{fig:kmeans-imputed-mother} and~\ref{fig:kmeans-imputed-kinase} present the corresponding K-Means analyses after the missing-data treatment described in Case II.

\subsection{Hierarchical Clustering}

\subsubsection{Complete Data Analysis}
\begin{figure}[H]
\centering
\includegraphics[width=0.48\textwidth]{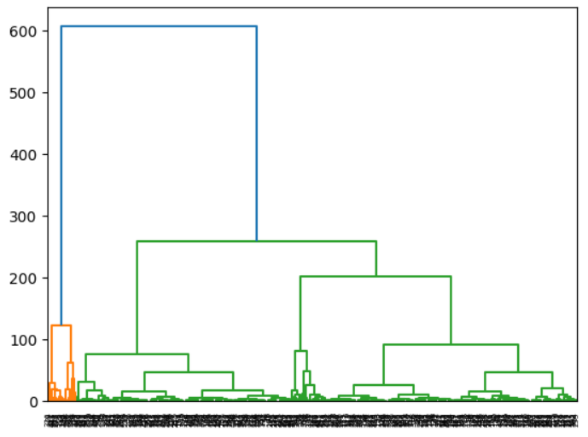}\hfill
\includegraphics[width=0.48\textwidth]{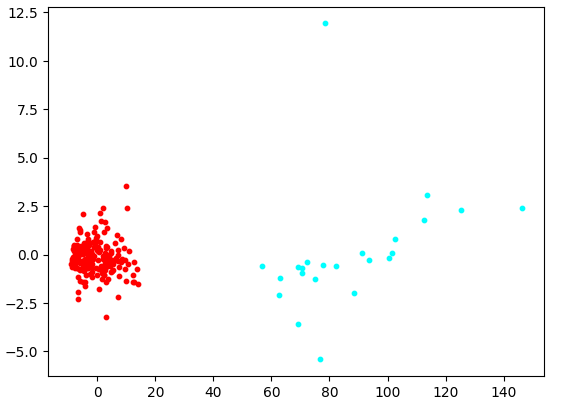}
\caption{Hierarchical clustering for the mother dataset under complete data analysis: dendrogram and clustered data.}
\label{fig:hier-mother}
\end{figure}

\begin{figure}[H]
\centering
\includegraphics[width=0.48\textwidth]{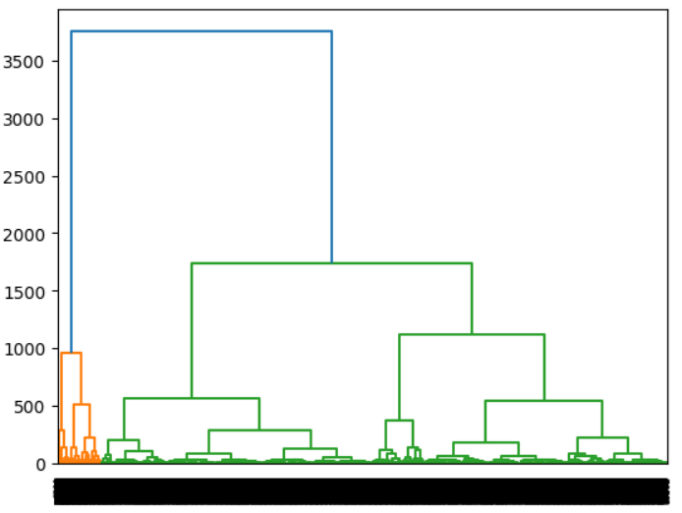}\hfill
\includegraphics[width=0.48\textwidth]{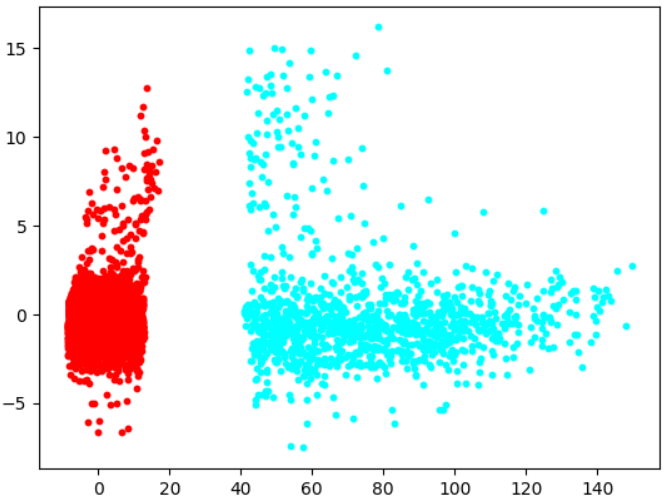}
\caption{Hierarchical clustering for the kinase dataset under complete data analysis: dendrogram and clustered data.}
\label{fig:hier-kinase}
\end{figure}

The hierarchical clustering results for the mother and kinase datasets are shown in Figures~\ref{fig:hier-mother} and~\ref{fig:hier-kinase}, respectively. Each pair contains the dendrogram and the corresponding clustered-data visualization.

\subsubsection{20\% Missingness Followed by Column-Mean Imputation}
\begin{figure}[H]
\centering
\includegraphics[width=0.48\textwidth]{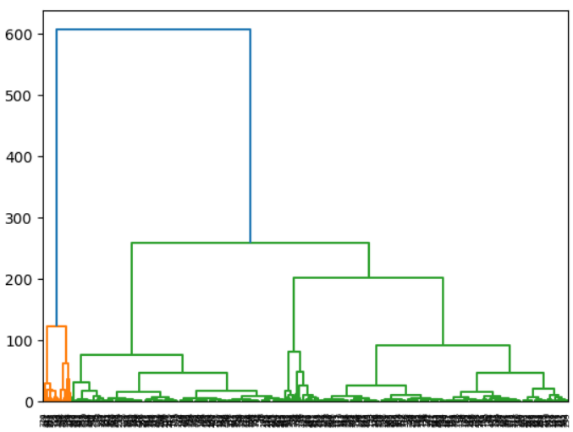}\hfill
\includegraphics[width=0.48\textwidth]{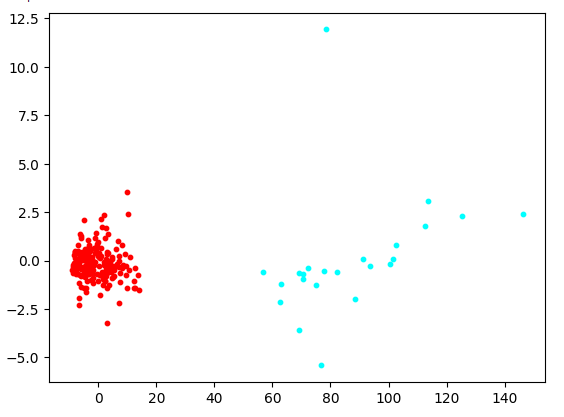}
\caption{Hierarchical clustering for the mother dataset under Case II.}
\label{fig:hier-imputed-mother}
\end{figure}

\begin{figure}[H]
\centering
\includegraphics[width=0.48\textwidth]{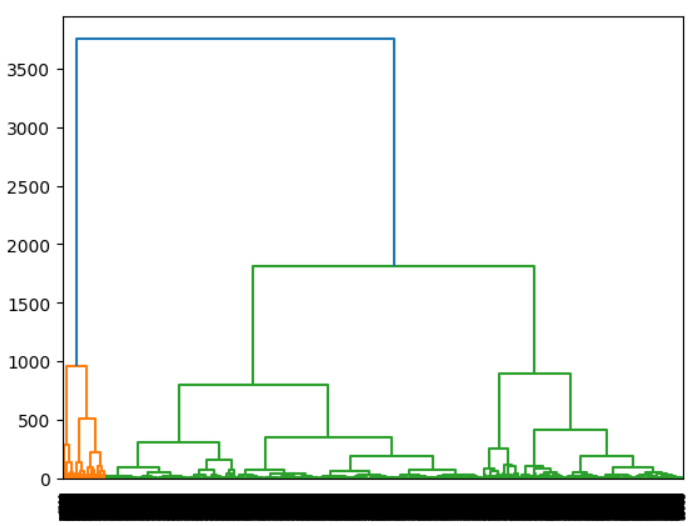}\hfill
\includegraphics[width=0.48\textwidth]{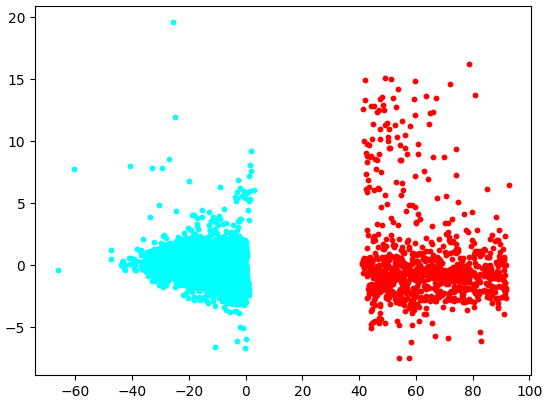}
\caption{Hierarchical clustering for the kinase dataset under Case II.}
\label{fig:hier-imputed-kinase}
\end{figure}

Figures~\ref{fig:hier-imputed-mother} and~\ref{fig:hier-imputed-kinase} show the hierarchical clustering results after mean imputation.

\subsection{Fuzzy C-Means Clustering}

\subsubsection{Complete Data Analysis}
\begin{figure}[H]
\centering
\includegraphics[width=0.48\textwidth]{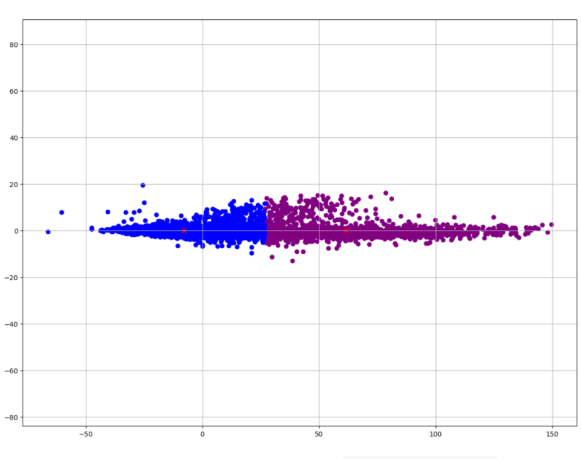}\hfill
\includegraphics[width=0.48\textwidth]{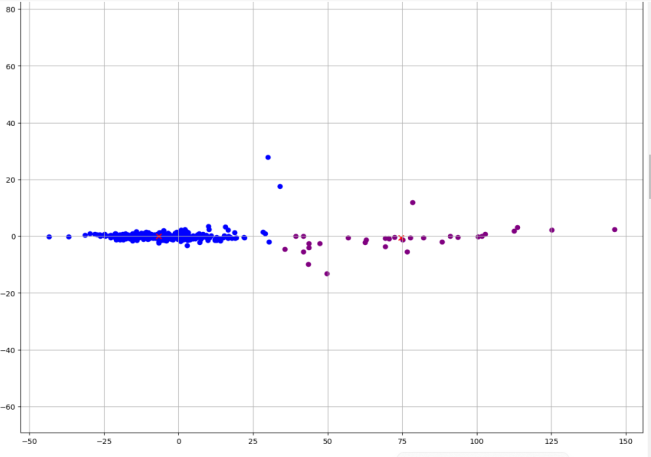}
\caption{Fuzzy C-Means clustering under complete data analysis for the mother and kinase datasets.}
\label{fig:fuzzy-complete}
\end{figure}

The Fuzzy C-Means visualizations for the mother and kinase datasets are shown in Figure~\ref{fig:fuzzy-complete}.

\subsubsection{20\% Missingness Followed by Column-Mean Imputation}
\begin{figure}[H]
\centering
\includegraphics[width=0.46\textwidth]{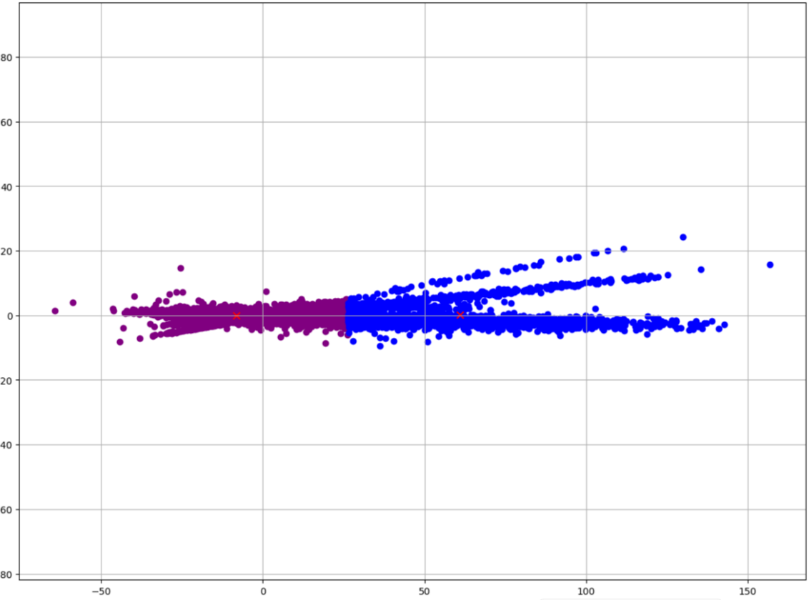}\hfill
\includegraphics[width=0.50\textwidth]{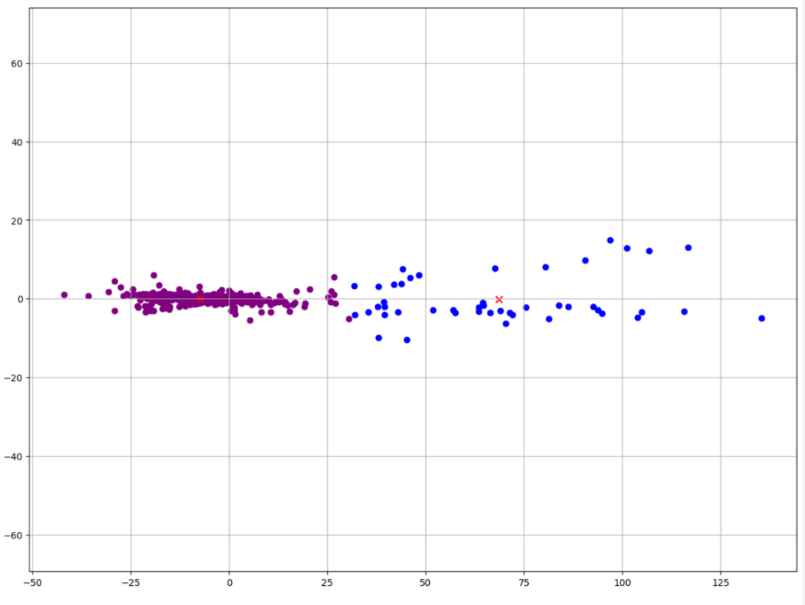}
\caption{Fuzzy C-Means clustering under Case II for the mother and kinase datasets.}
\label{fig:fuzzy-imputed}
\end{figure}

The corresponding Case II visualizations are presented in Figure~\ref{fig:fuzzy-imputed}.

\section{Discussion}
The clustering analyses provide consistent evidence that the human gene score data and the selected kinase data exhibit a two-cluster structure under the common set of 769 features. This result is observed for both the complete-data analysis and the mean-imputation analysis.

The silhouette scores remain relatively high across the different clustering methods. For the complete-data analysis, the silhouette scores for the gene score dataset range from 0.633 to 0.788, while those for the kinase dataset range from 0.724 to 0.825. Under mean imputation, the corresponding ranges are 0.633--0.788 for the gene score data and 0.795--0.825 for the kinase data.

\section{Conclusion and Future Scope}
The data analysis gives evidence that both the human gene score data and the kinase data are clustered into the same number of clusters under similar features. Across the two missing-data treatments and the four clustering methods considered, the optimal number of clusters was consistently two.

Further analysis can investigate the clustering structure under different feature combinations and alternative strategies for handling missing values. Such extensions may help determine whether the observed two-cluster structure remains stable when the feature space or data-preprocessing strategy is changed.

\section{Acknowledgements}

We acknowledge Dr. N. S. Sinha, School of Biological Sciences, Indian Institute of Science Education and Research (IISER) Thiruvananthapuram, for providing the human gene-score data used in this study.

\end{document}